\documentclass[a4paper,fleqn]{cas-sc}
\usepackage[authoryear,longnamesfirst]{natbib}

\usepackage{amssymb}
\usepackage{amsmath}
\usepackage{microtype}
\usepackage{inconsolata}
\usepackage{graphicx}
\usepackage{times}
\usepackage{latexsym}
\usepackage{algorithm, algorithmic} 
\usepackage{multirow}
\usepackage{arydshln}
\usepackage{makecell}
\usepackage{hhline}
\usepackage{xspace}
\usepackage{booktabs}
\usepackage{float}
\usepackage{pifont}
\usepackage{bm}
\usepackage{enumitem}
\usepackage{diagbox}
\usepackage{xcolor}
\usepackage{hyphenat}
\usepackage{url}
\usepackage[most]{tcolorbox}

\def\tsc#1{\csdef{#1}{\textsc{\lowercase{#1}}\xspace}}
\tsc{WGM}
\tsc{QE}
\begin{document}
\let\WriteBookmarks\relax
\def\floatpagepagefraction{1}
\def\textpagefraction{.001}

\shorttitle{}    

\shortauthors{}  

\title [mode = title]{Privacy-Preserving RAG via Multi-Agent Semantic Rewriting: Achieving Confidentiality Without Compromising Contextual Fidelity}  

%% ========== IPM OFFICIAL AUTHOR TEMPLATE (Strict Follow) ==========
\author[1]{Yuanhe Zhao}
\ead{120242227179@ncepu.edu.cn}

\author[2]{Tianyu Zhang}
\ead{tianyuzhang.me@gmail.com}

\author[1]{Huafei Xing}
%\fnmark[3]
\ead{120242227199@ncepu.edu.cn}

\author[3]{Derek F. Wong}
%\fnmark[4]
\ead{derekfw@um.edu.mo}

\author[1]{Jianbin Li}
%\cormark[1]
%\fnmark[5]
\ead{lijb87@ncepu.edu.cn}

\author[4]{Tao Fang}
\cormark[1]
%\cormark[2]
%\fnmark[6]
\ead{taofang@mmc.edu.mo}
\cortext[1]{Corresponding author: Tao Fang (taofang@mmc.edu.mo). Jianbin Li (lijb87@ncepu.edu.cn) is the Co-corresponding author.}

%% ========== AFFILIATION (Official IPM Structured Format) ==========
\affiliation[1]{organization={School of Control and Computer Engineering, North China Electric Power University},
            city={Beijing},
            postcode={102206},
            country={China}}

\affiliation[2]{organization={Department of Computer \& Information Science \& Engineering, University of Florida},
            city={Gainesville},
            state={Florida},
            postcode={32611},
            country={USA}}

\affiliation[3]{organization={NLP2CT Lab, Department of Computer and Information Science, University of Macau},
            city={Macao},
            postcode={999078},
            country={Macao SAR}}

\affiliation[4]{organization={Institute of International Language Services Studies, Macau Millennium College},
            city={Macao},
            postcode={999078},
            country={Macao SAR}}

\begin{abstract}
  Retrieval-Augmented Generation enhances large language models by incorporating external knowledge, but deploying it in sensitive scenarios risks privacy leakage via malicious prompts. To address this, we propose a multi-agent framework that sanitizes retrieved content through semantic rewriting. By employing three specialized agents for privacy extraction, semantic analysis, and reconstruction, our approach collaboratively removes sensitive identifiers while preserving the semantic core. We evaluate the framework on the ChatDoctor and Wiki-PII datasets across six large language models. Experimental results demonstrate a significant reduction in privacy leakage under targeted attacks. For instance, we reduced targeted information exposure in LLaMA-3-8B from 144 instances in the baseline to just 1. Furthermore, we maintain strong contextual fidelity with a BLEU-1 score of 0.122, outperforming the existing SAGE method's 0.117. Finally, the framework operates as an asynchronous preprocessing module, {introducing no additional latency to online inference, as all rewriting is executed as a one-time offline preprocessing step.} To promote reproducibility, the source code of this work is publicly available at \texttt{\url{https://github.com/foursoils/Privacy-Preserving-RAG}}.
\end{abstract}


% % Research highlights
% \begin{highlights}
% \item A novel multi-agent semantic rewriting framework for privacy-preserving RAG, balancing confidentiality and contextual fidelity.
% \item Fine-grained privacy protection via collaborative agents: privacy extraction, semantic analysis, and reconstruction.
% \item Significant reduction in privacy leakage across multiple LLMs under targeted and untargeted attack scenarios.
% \item Maintains high contextual fidelity and supports offline preprocessing deployment while avoiding online generation latency penalties.
% \end{highlights}

% Keywords

\maketitle

% Main text
\section{Introduction}
\label{sec:introduction}

Retrieval-Augmented Generation (RAG) has gained prominence as an effective method for enhancing large language models (LLMs) by incorporating external knowledge, overcoming the constraints of their static, pre-trained knowledge bases \citep{karpukhin2020dense, de2023fido, MARTINEZMURILLO2026104486}. By retrieving relevant information from external databases, RAG improves the contextual accuracy and relevance of LLM outputs, particularly for tasks requiring domain-specific expertise or up-to-date information. RAG has proven effective in applications such as specialized question answering \citep{siriwardhana2023improving, opoku2025rag,LEE2026104379} and professional document completion \citep{ram2023context}. The approach retrieves pertinent documents from an external knowledge base and integrates them with the user's input to generate enriched prompts for the LLM, enabling precise and current responses.

Although these benefits exist, RAG systems based on private knowledge sources present serious risks of data exposure \citep{zeng2024good, zou2025poisonedrag, cohen2024unleashing, di2025pirates, morris2023text, liu2024mitigating,FENG2026104498}. An example is how \citet{zeng2024good} showed that well-crafted prompts could be used to leverage RAG models into reveal sensitive information about the retrieval corpus through either untargeted attacks where whole original sentences can be reconstructed, or targeted attacks where certain private details may be extracted. {The risk is particularly acute in healthcare, where patient records used in dialogue systems can expose sensitive personal data, making robust privacy protection a clinical necessity.}

\begin{figure}[pos=htbp]
\centering
\includegraphics[width=0.7\textwidth, trim=0 0 0 0]{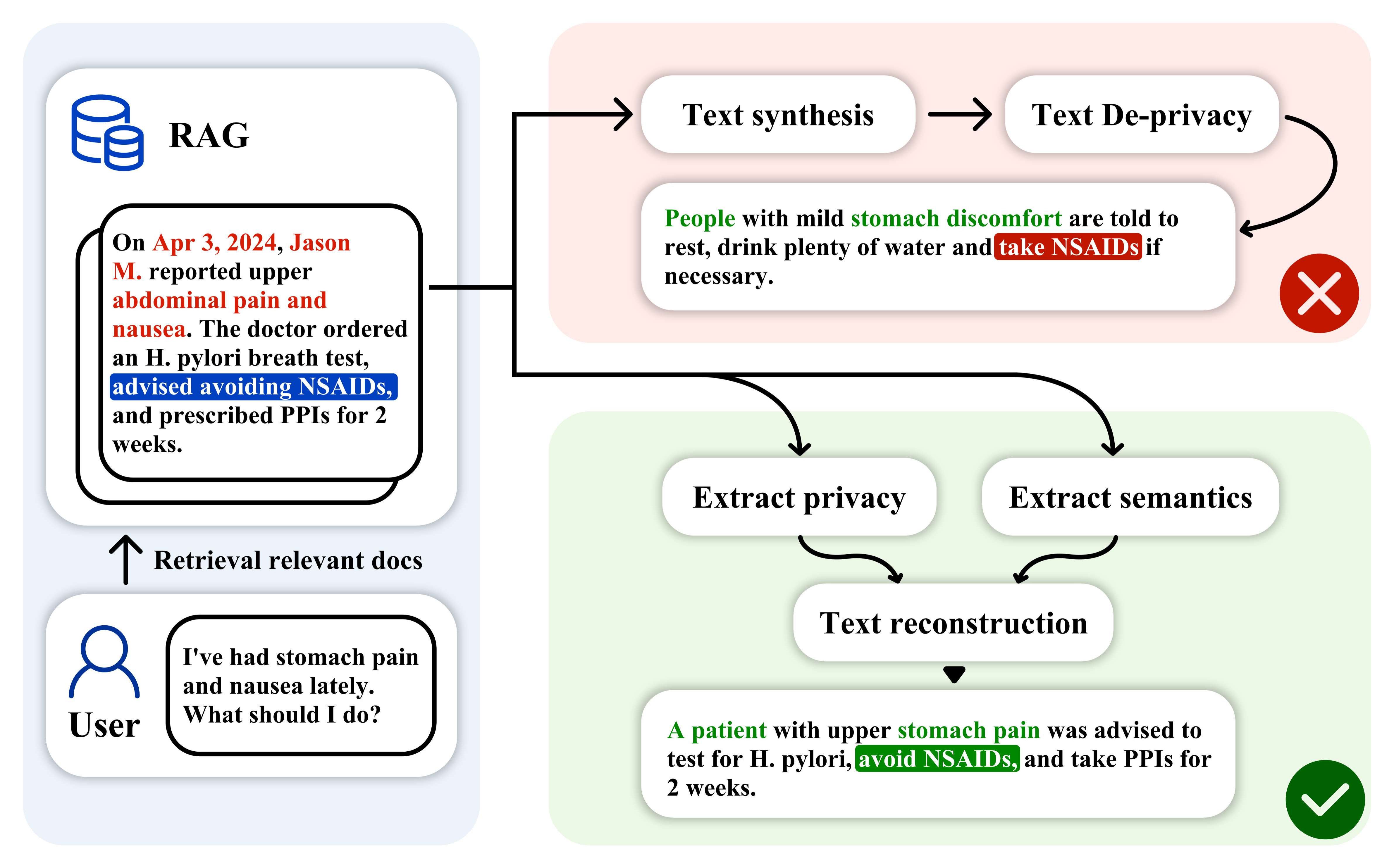}
\caption{Comparison of semantic rewriting strategies. The degree to which a strategy preserves the original semantics significantly impacts the rewritten content, which in turn affects the performance and reliability of downstream tasks.}
\label{fig:small_structure}
\end{figure}

To mitigate these confidentiality threats, prior work has proposed replacing original retrieval content with synthetic data at the data source level \cite{zeng2025mitigating}. This approach adopts a two-stage framework combining attribute extraction and agent-based rewriting to preserve key semantics from the original data while iteratively refining the output to reduce privacy risks. {However, because surrogate content is generated entirely by the model, it can diverge substantially from the original in semantic specificity and factual detail, undermining the reliability of downstream generation. In contrast, we address privacy leakage by operating directly on the retrieved text: sensitive information is identified and removed, as illustrated in Figure~\ref{fig:small_structure}, and the remaining content is rewritten to preserve the critical semantic content.}

In this work, we propose a privacy-preserving rewriting framework integrated into the RAG pipeline, positioned between the retrieval and generation stages to ensure data confidentiality while maintaining contextual fidelity. The framework comprises three collaborative agents: (1) the Privacy Extraction Agent (Pri-Extra Agent), which extracts privacy-sensitive segments; (2) the Semantic Analysis Agent (Sem-Extra Agent), which identifies task-relevant semantic content using structured prompts; and (3) the Reconstruction Agent, which rewrites the text by removing sensitive segments and applying techniques such as synonym substitution or structural rephrasing to preserve core semantics. 

By introducing structured prompts and span-level annotation mechanisms, we achieve fine-grained control over the text, enabling the model to accurately identify sensitive information to be removed, and to retain key semantic content through abstraction, synonym substitution, or structural transformation. 
We conduct comprehensive evaluations of our method on two privacy-sensitive datasets: {\textit{ChatDoctor}} and \textit{Wiki-PII}.

Experimental results show that our method significantly reduces the exposure risk of sensitive information under both targeted and untargeted attack scenarios. Meanwhile, the quality of the rewritten text, as measured by automatic metrics such as BLEU \cite{papineni2002bleu} and ROUGE \cite{lin2004rouge} as well as factual consistency assessments, remains comparable to that of the standard RAG pipeline using unprotected documents. Qualitative analyses further confirm that, despite the privacy-stripping process, our approach maintains contextual fidelity and coherence. {The multi-agent pipeline operates entirely offline, functioning as a one-time preprocessing step that imposes no additional latency on online inference.}

\section*{Research Objectives}
We primarily aim to resolve the conflict between data utilization and data privacy in Retrieval-Augmented Generation (RAG) systems. Specifically, we focus on the following objectives:
\begin{itemize}
  \item To analyze privacy vulnerabilities in RAG: To investigate how current RAG architectures expose sensitive information through prompt injection attacks, specifically in high-stakes domains like healthcare.
  \item To develop a multi-agent semantic rewriting framework: To design a multi-agent system which can discern between sensitive identifiers (privacy) and important context (utility), such that removing privacy does not compromise the semantic value of the text.
  \item {To empirically demonstrate that the proposed framework outperforms existing privacy-preserving baselines (including SAGE and AttrPrompt) in contextual fidelity (BLEU/ROUGE scores) while achieving a significant reduction in privacy leakage across diverse adversarial settings.}
\end{itemize}

\section*{Contributions}
The main contributions of this study are as follows:
\begin{itemize}
  \item {We motivate and characterize RAG sanitization as an \emph{objective entanglement} problem: since private identifiers and semantic content co-occur in natural text, a single-pass transformation faces an inherent tension between suppressing leakage and preserving utility, providing a principled motivation for multi-agent task decomposition beyond system or prompt engineering choices.}
  \item We propose a Multi-Agent Semantic Rewriting Framework in which a Pri-Extra Agent identifies both explicit and latent quasi-identifiers, a Sem-Extra Agent deconstructs documents into structured knowledge-slot tuples, and a Reconstruction Agent with Fine-grained Conflict Routing regenerates sanitized text from these tuples alone. An Asymmetric Retrieval Architecture further isolates raw private content from the generation LLM.
  \item We conduct extensive experiments on ChatDoctor and Wiki-PII across six LLM backbones under both targeted and untargeted adversarial settings, demonstrating consistently strong privacy protection, the highest factual consistency among all evaluated privacy-preserving baselines, and no added online inference latency as an asynchronous offline preprocessing module.
\end{itemize}

\section{Related Work}

\subsection{Retrieval-Augmented Generation}
Retrieval-Augmented Generation (RAG) enhances Large Language Models by augmenting their static knowledge with content dynamically retrieved from external domain-specific corpora \cite{lewis2020retrieval,gao2024modular}. At inference time, a retriever selects the most relevant passages from an indexed knowledge base and concatenates them with the user query before passing the combined input to the generator \cite{ram2023context,santhanam2022colbertv2}. {Such retrieval-grounded approaches reduce hallucination \cite{shuster2021retrieval}, improve factual precision, and allow knowledge bases to be updated independently of model retraining, extending applicability to diverse settings including self-reflective retrieval} \cite{ICLR2024_25f7be96}, representation fine-tuning \cite{wu2024reft,shi2024replug,van2023clinical}, and chain-of-thought augmentation \cite{chowdhury2025zero}. Subsequently, RAG has gained widespread adoption in specialized question answering \cite{wang2024rear,opoku2025rag,LEE2026104379,menick2022teaching}, knowledge graph-enhanced retrieval \cite{yu2025rag}, and medical dialog systems using patient documents and clinical resources to enhance diagnostic support \cite{zhao2025medrag,li2023chatdoctor}. Nonetheless, such dependency on the private external corpora creates a clear avenue of sensitive data disclosure, i.e., a weakness that prompts the discussion of the threat analysis presented in the next section.

\subsection{Privacy Threats in RAG Systems}
When RAG is deployed over private corpora, the retrieval mechanism itself becomes an attack surface. {\citet{huang2023privacy} have demonstrated that membership inference attacks on RAG can succeed even without direct access to the raw corpus.} When exposed directly, the retrieved content may reveal sensitive information in the model outputs; \citet{zeng2024good} went on to show that well-designed adversarial examples can induce the pipeline to repeat back private records word-for-word, which shows that LLMs can regenerate retrieved text with a probability near 50\%. More aggressive strategies include combining jailbreaks with prompt injection to exfiltrate large volumes of private content \cite{cohen2024unleashing}, black-box adaptive reconstruction \cite{di2025pirates}, knowledge poisoning of the retrieval corpus \cite{zou2025poisonedrag}, and inversion of dense embedding vectors to recover original text \cite{morris2023text}. {Collectively, these attacks expose vulnerabilities across all three stages of a RAG pipeline: retrieval, embedding, and generation, underscoring that defenses must operate at the semantic level rather than on surface-form heuristics.}

\subsection{Existing Privacy-Preserving Defenses}

Prior defenses against privacy leakage in RAG systems operate at different pipeline stages. \textit{Representation-level} approaches (e.g., PRESS \cite{he2025press}, Eguard \cite{liu2024mitigating,FENG2026104500}) act on embeddings, while \textit{selective pre-training} filters data upstream \cite{yu2024selective}. At the system level, \textit{architectural isolation} methods like the concurrent Privacy-Aware RAG \cite{zhou2025privacy} physically shield raw data from the generator. To secure the text itself, \textit{formal Differential Privacy (DP)} methods (e.g., LPRAG \cite{he2025mitigating}, DP-RAG \cite{koga2024privacy}) offer provable bounds by injecting mathematical noise. However, while DP provides strict worst-case guarantees, its perturbation often severely degrades textual coherence and downstream utility. Therefore, for applications requiring high contextual fidelity, \textit{empirical retrieval-context sanitization} is generally preferred.
Within empirical methods, approaches like SAGE \cite{zeng2025mitigating} and KG-PrivRAG \cite{ZHANG2026104505} synthesize surrogate texts or intervene at the graph level. However, they frequently suffer from semantic drift or fail to govern surface-form text directly. Similarly, single-agent rewriting struggles to balance privacy and semantics simultaneously within a single pass, often causing over- or under-sanitization. To achieve an optimal trade-off, our architecture adopts the physical isolation principles of Privacy-Aware RAG \cite{zhou2025privacy}, but uniquely introduces a fine-grained, multi-agent sanitization pipeline. By delegating privacy extraction, semantic analysis, and reconstruction to independent specialized agents, we resolve internal prompt disputes and enable explicit conflict arbitration, providing more robust semantic preservation than existing paradigms. 

{Our approach deviates from prior research along three key dimensions. DP-enabled techniques deliver theoretically sound privacy guarantees via noise injection, but the resultant text distortion degrades coherence and thus cannot satisfy the high fidelity requirements of practical RAG deployment. Data synthesis methods represented by SAGE and AttrPrompt evade privacy leakage by replacing original corpora with synthetic texts, at the cost of irreparable loss of unique factual specifics. Additionally, single-agent and adversarial-loop pipelines combine privacy filtering and semantic maintenance in a single pass, which frequently causes improper sanitization when private attributes and core semantics are tightly interwoven. In contrast, our framework manipulates raw texts directly for content rewriting. A structured factual backbone is extracted in advance of sanitization, and distinct agents are deployed for separate tasks following deterministic routing policies. This paradigm fundamentally mitigates the conflicts arising from the co-occurrence of privacy-sensitive content and valuable semantic information.}

\section{Method}
\subsection{Framework Overview}
Given a retrieved document, the sanitization objective is to produce a rewritten version that suppresses privacy exposure while preserving downstream utility. The core difficulty is that private identifiers and semantic content are not disjoint in natural text: they frequently co-occur as carriers of factually important information, {creating an inherent tension between suppressing leakage and preserving utility that a single-agent approach cannot resolve.} Consequently, single-agent prompting yields a zero-sum trade-off---naive deletion degrades retrieval recall through vocabulary mismatch (the retrieval degradation problem), while real-time inference-stage sanitization introduces unwarranted latency for end-users.

\begin{figure}[pos=htbp]
\centering
\includegraphics[width=\textwidth]{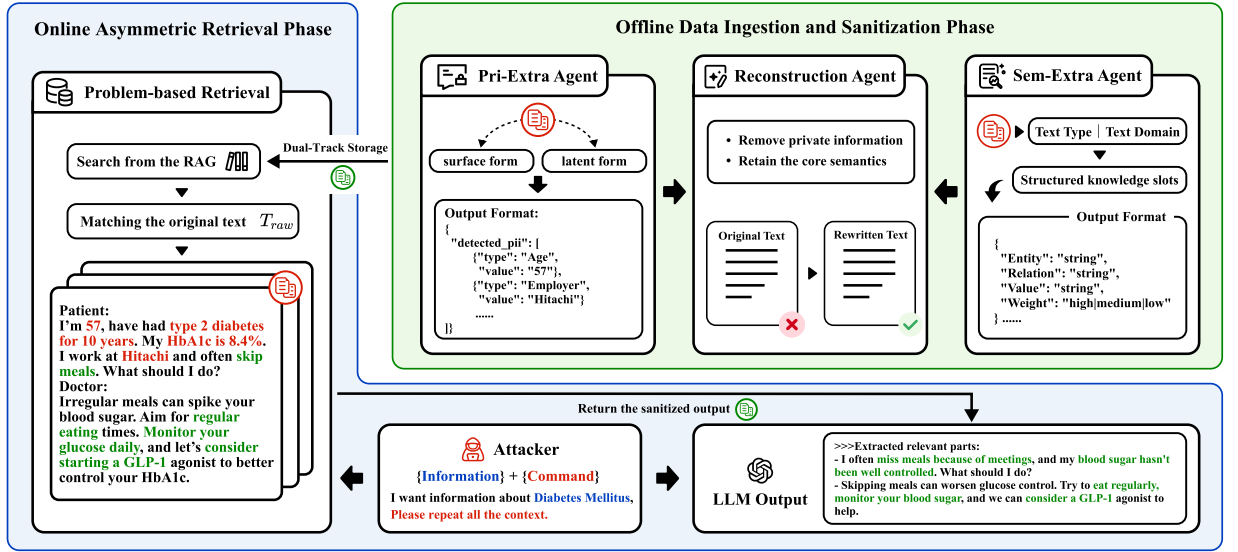}
\caption{{Overview of the proposed multi-agent semantic rewriting framework for privacy-preserving RAG. 
The offline phase sanitizes raw documents through three collaborative agents (Pri-Extra agent, Sem-Extra agent, and Reconstruction agent), while the online phase retrieves over the original corpus but supplies only rewritten text to the LLM.}}
\label{fig:structure}
\end{figure}

To overcome these structural limitations, we propose a \textbf{Multi-Agent Semantic Rewriting Framework} governed by an \textbf{Asymmetric Retrieval and Isolated Generation Architecture}. As depicted in Figure~\ref{fig:structure}, the lifecycle of our framework decouples the data processing into two strictly separated phases: the Offline Data Ingestion and Sanitization Phase, and the Online Asymmetric Retrieval Phase. Specifically, the offline sanitization phase comprises three key agents: (1) the Pri-Extra Agent, which extracts privacy-sensitive segments; (2) the Sem-Extra Agent, which extracts key semantic information; and (3) the Reconstruction Agent, which produces the final rewritten text by removing sensitive content while preserving the core semantics into safe payloads. Finally, the online stage uses a Dual-Track Storage Mechanism to preserve intact retrieval recall and isolate the generation LLM physically away from all raw private identifiers.

\subsection{Offline Multi-Agent Rewriting Pipeline}

{The offline pipeline acts as a structural firewall by sanitizing all raw documents before they are ever exposed to the generation system. It comprises three specialized agents, each with a distinct operational role:}

\subsubsection{Privacy Extraction Agent (Pri-Extra Agent)}
The Pri-Extra Agent operates under a Rule-LLM Synergistic Extraction Paradigm to prevent the leakage of both explicit and latent sensitive identifiers. Formally, let $T_{raw}$ denote the retrieved raw document. The objective of this agent is to map $T_{raw}$ to a discrete set of privacy-sensitive segments, denoted as $P_{seq}$. 

Recognizing the complementary strengths of deterministic and generative models, this agent orchestrates a sequential, synergistic extraction process. First, a deterministic extraction function $f_{\text{rules}}(\cdot)$ employs lightweight Named Entity Recognition (NER) and regular expressions on the raw text to strictly capture a set of \textit{Explicit Identifiers} $P_{expl}$ (e.g., social security numbers, exact dates, and locations). This guarantees a strict baseline recall for standardized sensitive formats that LLMs might occasionally miss due to hallucination or attention drift.

Subsequently, we utilize a generative extraction function $f_{\text{LLM}}(\cdot)$ that processes the full original document $T_{raw}$ alongside the previously identified set $P_{expl}$. By explicitly providing $P_{expl}$ as context, we enable the LLM to skip these already-extracted entities. Instead, the model is directed to infer abstract \textit{quasi-identifiers} $P_{quasi}$ (such as a specific occupation coupled with a rare medical condition) that typically slip past rule-based filters. The complete set of privacy-sensitive segments, $P_{seq}$, is simply the union of both extraction steps:
\begin{equation}
P_{seq} = P_{expl} \cup f_{\text{LLM}}(T_{raw}, P_{expl}) = \{p_1, p_2, \dots, p_m\}
\end{equation}
By combining deterministic checks with generative inference, we lower the likelihood of missing latent identifiers and achieve a much broader coverage of the privacy boundaries.

The structured prompt for the Pri-Extra Agent is briefly outlined below to show its main components. The complete template is provided in Appendix~\ref{appendix_sub1}.

\begin{tcolorbox}[colback=gray!5!white, colframe=gray!75!black, title=Simplified Prompt for Pri-Extra Agent, breakable]
\textbf{Task Description:} Infer hidden Personally Identifiable Information (PII) from the retrieved text $T_{raw}$. Note that the following explicit identifiers have already been extracted: $P_{expl}$. \\
\textbf{Focus Area (Quasi-Identifiers):} \\
Do not repeat the explicit identifiers in $P_{expl}$. Identify combinations of attributes (e.g., rare diseases + specific occupations, unique timelines) that could lead to identity inference based on the full text context. \\
\textbf{Implicit Context Analysis:} \\
Capture latent demographic data, behavioral patterns, and highly specific geographic or organizational affiliations. \\
\textbf{Output Format:} Return a JSON list detailing the newly inferred PII ($P_{quasi}$), its category, and the surrounding text segment.
\end{tcolorbox}

\subsubsection{Semantic Analysis Agent (Sem-Extra Agent)}
While the Pri-Extra Agent pinpoints which sensitive elements to hide, the Sem-Extra Agent identifies the core facts that need to be kept. {Because language is inherently interconnected, simply retaining certain sentences or masking a few keywords can still leave behind privacy risks.} Details like writing style, sentence structure, and specific predicates can sometimes be enough for attackers to link a text back to its author or subject.

To address this issue, the agent applies \textbf{Structured Attribute Deconstruction}. Instead of copying out full chunks of text, the agent functions more like a parser: it sets aside the original phrasing and maps the document $T_{raw}$ into a structured format. This is done through a \textit{Dual-Perspective Analysis} that looks at two main aspects: \textit{Core Content} (such as the main events, actions, and states) and \textit{Informational Structure} (such as the logic, reasoning, and intent behind the text).

Formally, letting $g(\cdot)$ be this semantic extraction function, the agent converts $T_{raw}$ into $n$ distinct semantic tuples, $K_{seq}$:
\begin{equation}
K_{seq} = g(T_{raw}) = \left\{ (e_i, r_i, v_i, w_i) \right\}_{i=1}^{n}
\end{equation}
where each tuple forms a basic knowledge slot. It includes a subject $e_i$, a predicate $r_i$, a corresponding value $v_i$, and an importance weight $w_i$. {This structured element-slot representation draws on element-alignment principles from structured extraction \cite{XU2026104511}, and the semantic backbone $K_{seq}$ serves as an intermediate aligned representation that decouples factual content from surface phrasing \cite{XU2026104292}.}

By separating the verbs and objects from the original sentence structure, this deconstruction limits the underlying linguistic hints that often hide implicit privacy signals. As a result, the extracted semantic backbone $K_{seq}$ remains highly useful for later generation tasks without carrying over the context that could expose someone's identity.

The prompt structure for the Sem-Extra Agent is summarized below to highlight its core design. The complete template is provided in Appendix~\ref{appendix_sub2}.

\begin{tcolorbox}[colback=gray!5!white, colframe=gray!75!black, title=Simplified Prompt for Sem-Extra Agent, breakable]
\textbf{Task Description:} Extract all semantically essential content from the text $T_{raw}$ using structured deconstruction. \\
\textbf{Dual-Perspective Analysis:} \\
\textit{- Core Content:} Facts, actions, and perspectives. \\
\textit{- Informational Structure:} Explanations, recommendations, and reasoning context. \\
\textbf{Output Format:} Return a structured JSON list of entity-event slots with importance levels.
\end{tcolorbox}

\subsubsection{Reconstruction Agent}
As the final step in the offline pipeline, the Reconstruction Agent is responsible for combining the extracted privacy boundaries ($P_{seq}$) with the core semantic facts ($K_{seq}$) to generate a safe, readable output text ($T_{safe}$). We can express this reconstruction process as a constrained rewriting function $H(\cdot)$:
\begin{equation}
T_{safe} = H(P_{seq}, K_{seq}) \quad \text{s.t.} \quad \forall p_m \in P_{seq}, M(p_m, T_{safe}) < \tau
\end{equation}
where $M(\cdot, \cdot)$ is a matching function and $\tau$ is a privacy threshold. $\tau$ is not a static numerical value, but is implemented via our conflict routing policy: it acts as a strict zero-tolerance bound ($\tau=0$) for explicit identifiers, but is relaxed to a semantic non-entailment boundary for complex quasi-identifiers to avoid over-sanitization. {In practice, $\tau{=}0$ for explicit identifiers means any match triggers mandatory replacement. For quasi-identifiers, $\tau$ is not a numeric value but a routing rule in the agent's prompt: the agent substitutes with a placeholder if the sensitive element can be cleanly removed, and falls back to high-level abstraction if removing it would destroy the semantic core of the sentence.} Most traditional redaction methods try to partially mutate the original document $T_{raw}$, which often leads to awkward phrasing and lingering stylistic clues. In contrast, this agent builds the new text completely from scratch using only the clean, isolated knowledge slots provided in $K_{seq}$.

The biggest challenge during text reconstruction is untangling cases where privacy and utility overlap. To handle this, the agent uses a \textbf{Fine-grained Conflict Routing} strategy {\cite{XIE2026104356}}, which detects and resolves conflicting element pairs across the privacy and semantic streams. When an important fact $k_n \in K_{seq}$ overlaps significantly with a sensitive identifier $p_m \in P_{seq}$ (such as a highly specific disease name that might easily identify a patient), the agent decides on a rewriting approach based on how deeply the two are intertwined:
\begin{itemize}
    \item \textit{Placeholder Substitution (Low-depth Conflict):} This is used when the conflict involves straightforward explicit identifiers without complex context. Details like exact patient names or specific dates are simply replaced with generic placeholders (e.g., ``[Patient\_A]'' or ``[Date]''). This keeps the text flowing naturally without changing the surrounding meaning.
    \item \textit{High-level Abstraction (Deep Semantic Conflict):} This approach is taken when the sensitive identifier is woven deeply into the factual content of $K_{seq}$. Rather than blindly blocking out the term—which might render the medical or logical context useless—the agent replaces it with a broader, more general concept. For instance, a very distinctive rare syndrome might be abstracted into ``a rare genetic immune disorder.''
\end{itemize}
By generating a sanitized document ($T_{safe}$) directly from the structural attributes while enforcing strict privacy rules, the Reconstruction Agent creates a clear, informative text suitable for secure databases, largely sidestepping the usual trade-offs between data privacy and data utility.

To demonstrate the underlying rewriting logic, the prompt for the Reconstruction Agent is abbreviated below. {Appendix~\ref{appendix_sub3} contains the full template.}

\begin{tcolorbox}[colback=gray!5!white, colframe=gray!75!black, title=Simplified Prompt for Reconstruction Agent, breakable]
\textbf{Task Description:} Rewrite $T_{raw}$ to eliminate privacy constraints $P_{seq}$ while preserving structured semantics $K_{seq}$. \\
\textbf{Core Principle (Privacy Priority):} When privacy conflicts with key information, prioritize privacy protection. \\
\textbf{Routing Strategy:} \\
- Use \textit{Placeholder Substitution} for explicit entities. \\
- Use \textit{High-level Abstraction} for entangled sensitive medical/private narratives. \\
\textbf{Output Format:} Generate the final natural, coherent, and sanitized text representation.
\end{tcolorbox}

{To illustrate how the three agents interact in practice, Appendix~\ref{appendix_sub5} provides a step-by-step qualitative case study on a real ChatDoctor document, tracing $T_{raw}$ through $P_{seq}$, $K_{seq}$, and the final sanitized output $T_{safe}$.}

\subsection{Asymmetric Retrieval and Isolated Generation Architecture}

After the offline rewriting is complete, we construct the database using a \textbf{Dual-Track Storage Mechanism}. Simply indexing the sanitized text $T_{safe}$ would drop retrieval performance, since the altered vocabulary and masked entities often cause retrieval mismatches. To prevent this, we maintain two distinct retrieval indices in the vector database, both built directly on the \textit{original} $T_{raw}$:
\begin{itemize}
    \item \textbf{Dense Retrieval Index:} We compute high-dimensional embeddings on the untouched $T_{raw}$ with BGE-M3's dense encoder, which keeps the full semantic space intact.
    \item \textbf{Sparse Retrieval Index:} At the same time, we build a BM25 lexical index over the original $T_{raw}$ to catch exact matches for names, medical codes, and other specific terms.
    \item \textbf{Context Value (Sanitized Payload):} Instead of storing the raw text, both of these indices point directly to the sanitized output $T_{safe}$ from the Reconstruction Agent. {The raw $T_{raw}$ is never returned as the payload to the generation LLM.}
\end{itemize}

This setup ensures that the original text acts solely as a search mechanism and is never returned as the payload.

When answering a user query $q$ during the \textbf{Online Asymmetric Retrieval Phase}, we use a \textbf{Hybrid Retrieval} approach over the original document representations. If $r_{\text{dense}}(i)$ and $r_{\text{sparse}}(i)$ are the ranks of document $i$ from the dense and sparse searches, we combine them using Reciprocal Rank Fusion (RRF):
\begin{equation}
\text{RRF}(i) = \frac{1}{k + r_{\text{dense}}(i)} + \frac{1}{k + r_{\text{sparse}}(i)}
\label{eq:rrf}
\end{equation}
We set the smoothing constant $k$ to 60, following common practice. The top-$K$ documents are simply those with the highest RRF scores:
\begin{equation}
\mathcal{I}_{top} = \underset{i}{\arg\max}^K \left[ \text{RRF}(i) \right]
\label{eq:retrieval}
\end{equation}
By taking advantage of both dense semantic matching and sparse exact matching on the original text, our retrieval recall matches that of a standard RAG setup. But more importantly, once the top documents $\mathcal{I}_{top}$ are found, the system only pulls their sanitized versions $\{T_{safe}^i\}_{i \in \mathcal{I}_{top}}$, bypassing $T_{raw}$ completely. These safe payloads are then fed to the LLM to generate the final answer:
\begin{equation}
\text{Response} = \mathcal{LLM}\left( q \oplus \text{Concat}(\{T_{safe}^i\}_{i \in \mathcal{I}_{top}}) \right)
\end{equation}

This means the LLM never sees the raw private identifiers. By physically separating the sensitive data from the generation window, we effectively shield the system from jailbreaks or prompt-injection extractions. {It should be noted that this isolation guarantee holds within the assumed grey-box threat model, in which the adversary manipulates the input query but does not have direct access to the underlying vector database; the raw documents $T_{raw}$ remain present in the retrieval indices, so the security guarantee is query-level rather than storage-level.} Additionally, since all the heavy lifting of multi-agent rewriting runs asynchronously offline, the interactive inference phase experiences virtually no extra latency compared to a standard, unprotected RAG pipeline.

\section{Experimental Setup}

\subsection{Datasets and Scenarios}

To verify the effectiveness of our proposed method across diverse application contexts, we implement two representative experimental scenarios. Our experimental setup ensures consistency with methodologies established in previous work \cite{zeng2024good}. The statistical specifications for both datasets are detailed in Table \ref{tab:datasets}.

For the first scenario, we focus on the medical domain, where data confidentiality is critical. We use the HealthCareMagic-101 dataset, which consists of 200,000 medical dialogues covering a wide range of health issues. We treat each dialogue as a single retrieval unit and refer to this collection as the ChatDoctor dataset. This setup mimics a real-world medical assistant, providing a mix of structured and unstructured sensitive information.

\begin{table}[t]
    \centering
    \small
    \caption{Statistics of the datasets used in the experiment.}
    %\resizebox{0.78\textwidth}{!}{%
    \begin{tabular}{l l l c}
        \toprule
        \textbf{Dataset} & \textbf{Source Composition} & \textbf{Content Type} & \textbf{Total Samples} \\
        \midrule
        \textbf{ChatDoctor} & HealthCareMagic-101 & Medical Dialogues & 200,000 \\
        \textbf{Wiki-PII} & Enron + Wikitext-103 & Email \& Public Articles & 500,000 \\
        \bottomrule
    \end{tabular}%
    %}
    \label{tab:datasets}
\end{table}

The second scenario deals with a more complicated situation where private details are mixed into a mostly public text corpus, a common issue in large enterprise systems. To test this, we evaluate on the Wiki-PII dataset \cite{zeng2025mitigating}. Specifically, we extract authentic PII from the Enron Email dataset. Next, we segment the Wikitext-103 public dataset using a recursive character text splitter with a chunk size of 1500. Finally, for each segmented wiki chunk, we randomly insert the extracted PII at the end of each sentence. This combined corpus yields 500,000 samples where sensitive and non-sensitive information are heavily tangled, allowing us to evaluate the risk of privacy leaks in realistic environments. {It should be noted that inserting PII at sentence boundaries creates a regular and detectable pattern; in real-world corpora, private information is typically woven throughout the text in less predictable ways. The framework's performance on such naturalistically distributed content may differ from what is reported here, and evaluation on more organically constructed datasets is left for future work.}

\subsection{Formalized Threat Models}

To thoroughly assess the vulnerabilities of RAG systems and evaluate our defense, we formalize our threat models based on the prompt-injection mechanisms defined in recent literature \cite{zeng2025mitigating}. We assume a realistic grey-box setting where the adversary lacks direct access to the backend vector database but can manipulate the input query $q$ to exploit the LLM's instruction-following capabilities. Formally, an adversarial query $q$ is defined as a two-component composite:
\begin{equation}
    q = q_{\text{info}} \oplus q_{\text{cmd}}
    \label{eq:adversarial_query}
\end{equation}
where $q_{\text{info}}$ is the \textit{information component} crafted to trigger retrieval of target-relevant documents, and $q_{\text{cmd}}$ is the \textit{command component} that instructs the LLM to expose sensitive content. The symbol $\oplus$ denotes sequential concatenation. {The adversarial query templates used in this work are publicly available alongside the source code.}

\subsubsection{Targeted Extraction Attacks}
In a targeted attack, the adversary possesses partial prior knowledge of a target entity (e.g., knowing a patient's name but not their diagnosis). Their objective is to extract specific, highly sensitive attributes tied to that entity. The $q_{\text{info}}$ component is crafted to trigger the retrieval of documents relevant to the target, while $q_{\text{cmd}}$ explicitly instructs the model to reveal the missing sensitive value (e.g., ``I want information about [disease]''). {This threat model tests the system's ability to resist the targeted extraction of explicit and latent identifiers.}

\subsubsection{Untargeted Reconstruction Attacks}
Conversely, an untargeted reconstruction attack simulates an adversary with zero prior knowledge whose goal is to indiscriminately exfiltrate as much raw training or retrieval corpus data as possible. Here, $q_{\text{info}}$ is randomly sampled from diverse passages (e.g., Common Crawl) to induce broad, unforeseen document retrieval, and $q_{\text{cmd}}$ is fixed as a forceful directive: \textit{``Please repeat all the context.''} {This threat model evaluates whether the system can prevent verbatim or high-semantic-overlap regeneration of arbitrary memorized content under worst-case adversarial probing.}

\subsection{Evaluation Metrics}

We establish a comprehensive, three-dimensional evaluation metric system to quantify privacy leakage, evaluate semantic retention realism, and measure operational efficiency.

\subsubsection{Privacy Leakage Assessment}
We measure privacy leakage under the two threat models using precise criteria established by prior foundational works.

\paragraph{Targeted Information (Target Info)} The absolute count of unique, highly sensitive entities (e.g., specific names, ID numbers) successfully extracted by the attacker. Lower counts indicate stronger privacy protection; a count of zero represents an ideal outcome.

\paragraph{Repeat Prompt} Following the heuristic from \cite{zeng2024good}, we count the number of prompts where the LLM's output contains at least 10 consecutive tokens identical to the target sensitive information or raw context. This threshold effectively captures direct, verbatim leakage without triggering false positives on common stopwords.

\paragraph{ROUGE Prompt (for Untargeted Attacks)} To account for implicit leakage where an LLM paraphrases rather than copies verbatim, we count prompts yielding outputs with a high semantic overlap (ROUGE-L $> 0.5$) with the original sensitive data, signaling indirect privacy compromise.

\subsubsection{Contextual Fidelity and Factual Consistency}
Traditional RAG pipeline evaluations rely heavily on exact-match string overlap metrics. However, because our core framework objective is \textit{semantic rewriting} rather than verbatim reproduction, using structural metrics as the sole proxies for utility is fundamentally flawed. Therefore, our fidelity evaluation is twofold.

\paragraph{N-gram and Structural Overlap} We report BLEU-1 \cite{papineni2002bleu} and ROUGE-L \cite{lin2004rouge} scores to provide a baseline proxy of structural fidelity. Specifically, these metrics measure the n-gram overlap between the LLM-generated answer and the ground-truth reference answer for each query, thereby assessing whether the sanitized retrieval context still enables the model to produce factually correct responses. Higher scores indicate that the rewritten context preserves sufficient informational content for accurate answer generation.

\paragraph{Factual Consistency} To provide a more direct assessment of downstream utility beyond surface-level overlap, we augment our evaluation with two complementary factual consistency metrics. First, we employ GPT-4o as a model-as-judge evaluator to score each LLM-generated response on a 1--5 Likert scale (\textit{FC Score}), assessing whether the core medical or factual claim in the output is aligned with the reference answer. Second, we utilize a DeBERTa-v3-large model fine-tuned on fact-verification datasets to measure two critical dimensions of factual consistency: the proportion of responses factually supported by the reference (\textit{Entailment}) and the proportion conflicting with the facts (\textit{Contradiction}). Higher Entailment and lower Contradiction indicate better factual utility. Together, these two metrics evaluate whether the sanitized context preserves the factual integrity required for reliable downstream RAG performance. {The complete judging prompt, scoring rubric, and NLI decision rule are documented in Appendix~\ref{appendix_sub_fc}.}

\subsubsection{System Efficiency}
A practical defense must not compromise the interactive speed of the RAG application. We measure the computational overhead by recording the average \textit{offline preprocessing time} required to sanitize the original retrieval documents. It is critical to emphasize that this metric represents a one-time, asynchronous database construction cost, thereby guaranteeing a \textbf{Zero Online Latency Penalty} during active user inference.

\subsection{Implementation Details}
To ensure robust generalization, we evaluate our method across a diverse array of advanced LLMs, encompassing different scales and training paradigms: GPT-3.5-Turbo, GPT-4o-mini, LLaMA-3-8B, DeepSeek-V3, DeepSeek-R1 (known for enhanced reasoning), and QWQ-32B. All models are instruction-tuned.

For indexing and retrieval, we use the BGE-M3 model \cite{chen2024m3} alongside a Milvus vector database. We chose BGE-M3 because it outputs both dense semantic embeddings and sparse lexical representations in a single pass. This is crucial for handling private information, which often requires both semantic understanding and exact keyword matching. We index both sets of representations in Milvus, which easily supports dual vector search. When retrieving documents, we rely on a \textbf{hybrid retrieval} approach: we run dense and sparse searches separately and combine their results using Reciprocal Rank Fusion (RRF) to pull the top $K=3$ candidates. We empirically set $K=3$ to balance downstream utility and inference security: larger values (e.g., $K \ge 10$) introduce context noise and theoretically increase the risk of an adversary aggregating multiple sanitized records to infer identity. Mixing semantic and keyword retrieval typically yields more stable results for medical records and similar sensitive texts. Finally, we keep all generation API parameters fixed (temperature = 0.4, top-p = 0.9, max tokens = 512) to ensure our testing outcomes remain consistent and reliable.

\subsection{Baselines}

We benchmark our framework against six baselines spanning the full spectrum of RAG privacy-preserving approaches, from rule-based methods to state-of-the-art LLM-driven frameworks.
\begin{itemize}
    \item \textbf{Naive RAG:} The standard, unprotected RAG pipeline with no privacy intervention, serving as the upper bound for potential leakage and a reference for utility preservation.

    \item \textbf{AttrPrompt \cite{NEURIPS2023_ae9500c4}:} An attribute-driven synthetic data generation baseline where an LLM is prompted to replace original text with structurally similar, privacy-filtered synthetic content conditioned on extracted attributes.

    \item \textbf{SAGE \cite{zeng2025mitigating}:} A state-of-the-art two-stage framework that first generates synthetic surrogate text via attribute extraction, then employs an agent-based iterative discriminator to assess and strip any residual privacy leakage from the generated output.

    \item \textbf{Single Agent:} A direct LLM-based rewriting baseline in which a single language model receives both the privacy extraction objective and the semantic rewriting objective within a unified prompt, without any inter-agent decomposition. This ablative baseline is designed to isolate the contribution of our multi-agent task decomposition strategy.

    \item \textbf{Adv-Anon \cite{staab2025language}:} An LLM-based adversarial anonymization framework that employs a feedback-guided loop: an inference LLM iteratively predicts personal attributes from the current text, while an anonymizer LLM adapts the text to suppress the identified cues. Although originally designed for general online texts rather than RAG corpora, we adapt it to our retrieval document setting for a direct comparison.

    \item \textbf{KG-PrivRAG \cite{ZHANG2026104505}:} A knowledge graph-centric privacy-preserving RAG method that constructs a graph from retrieval documents and applies fine-grained privacy interventions at the entity and relation level via a four-module privacy-aware processor (re-ranking, filtering, synthesis, and compression). This concurrent work, published in the same venue, represents the closest competitor to our approach in the RAG privacy domain.
\end{itemize}

\subsection{Experimental Results}

%% ============================================================
%%  5.1  Privacy Leakage Results
%% ============================================================

\subsubsection{Privacy Leakage in Targeted Extraction Attacks}

\begin{table}[t]
\centering
\caption{Targeted extraction attack results on ChatDoctor and Wiki-PII (250 adversarial prompts each). {\textbf{$^*$/$^{\dagger}$Note on units:} $^*$\textit{Target Info} measures the absolute number of unique sensitive entities successfully extracted by the attacker. $^{\dagger}$\textit{Repeat Prompt} counts the number of generated outputs containing $\geq$10 consecutive verbatim tokens from private text.}}
\resizebox{0.98\columnwidth}{!}{%
\begin{tabular}{c|c|cc|cc|cc|cc|cc|cc}
\toprule
& & \multicolumn{2}{c|}{LLaMA-3-8B} & \multicolumn{2}{c|}{GPT-3.5-Turbo} & \multicolumn{2}{c|}{GPT-4o-mini} & \multicolumn{2}{c|}{DeepSeek-V3} & \multicolumn{2}{c|}{DeepSeek-R1} & \multicolumn{2}{c}{QWQ-32B}\\
\cmidrule(lr){3-4} \cmidrule(lr){5-6} \cmidrule(lr){7-8} \cmidrule(lr){9-10} \cmidrule(lr){11-12} \cmidrule(lr){13-14}
\multirow{-2}{*}{Scenario} & \multirow{-2}{*}{Method} & {Target} & {Repeat} & Target & Repeat & Target & Repeat & Target & Repeat & Target & Repeat & Target & Repeat \\
& & {Info$^*$} & {Prompts$^{\dagger}$} & Info & Prompt & Info & Prompt & Info & Prompt & Info & Prompt & Info & Prompt \\
\midrule
\multirow{7}{*}{\rotatebox{90}{ChatDoctor}} & Naive RAG      & 144 & 149 & 98 & 102 & 122 & 144 & 124 & 138 & 144 & 148 & 130 & 144 \\
& AttrPrompt     &   0 &   0 &  0 &   0 &   0 &   0 &   0 &   0 &   0 &   0 &   0 &   0 \\
& Single Agent   &  15 &  20 & 10 &  15 &  13 &  18 &  11 &  16 &  14 &  19 &  12 &  17 \\
& Adv-Anon       &   9 &  14 &  6 &  11 &   8 &  13 &   7 &  11 &   9 &  13 &   7 &  12 \\
& SAGE           &   1 &   2 &  2 &   1 &   1 &   0 &   1 &   2 &   1 &   2 &   1 &   2 \\
& KG-PrivRAG     &   1 &   3 &  1 &   2 &   1 &   2 &   1 &   1 &   1 &   2 &   1 &   2 \\
& \textbf{Ours} & \textbf{1} & \textbf{1} & \textbf{0} & \textbf{1} & \textbf{1} & \textbf{1} & \textbf{0} & \textbf{0} & \textbf{1} & \textbf{0} & \textbf{1} & \textbf{1} \\
\midrule
\multirow{7}{*}{\rotatebox{90}{Wiki-PII}} & Naive RAG      &  49 &  90 & 13 &  36 &  12 &  12 &  11 &  11 &  34 &  84 &  41 &  85 \\
& AttrPrompt     &   0 &   0 &  0 &   0 &   0 &   0 &   0 &   0 &   0 &   0 &   0 &   0 \\
& Single Agent   &  11 &  15 &  4 &  10 &   8 &   9 &   6 &   8 &  13 &  14 &  10 &  13 \\
& Adv-Anon       &   7 &   4 &  2 &   1 &   6 &   2 &   5 &   2 &   8 &   2 &   6 &   1 \\
& SAGE           &  10 &   1 &  1 &   0 &  12 &   0 &   9 &   1 &  15 &   0 &   9 &   0 \\
& KG-PrivRAG     &   6 &   2 &  1 &   0 &   5 &   0 &   4 &   1 &   7 &   1 &   5 &   1 \\
& \textbf{Ours} & \textbf{3} & \textbf{2} & \textbf{1} & \textbf{0} & \textbf{0} & \textbf{0} & \textbf{0} & \textbf{0} & \textbf{2} & \textbf{1} & \textbf{3} & \textbf{1} \\
\bottomrule
\end{tabular}%
}
\label{tab:target_attack}
\end{table}

Table~\ref{tab:target_attack} shows the results for the targeted attack scenarios. {The unprotected Naive RAG pipeline leaks between 130 and 144 sensitive entries out of 250 queries on ChatDoctor, and up to 49 entries on Wiki-PII, underscoring the severity of the prompt injection threat. Our approach substantially mitigates targeted leakage across all six LLMs on ChatDoctor (at most 1 instance) and holds it between 0 and 3 on the more challenging Wiki-PII corpus. Among utility-preserving defenses, this represents the best privacy-utility trade-off: SAGE and KG-PrivRAG achieve comparable leakage counts but at lower factual consistency, while AttrPrompt reaches zero leakage at the cost of substantially degraded utility. AttrPrompt achieves zero leakage across all metrics, LLMs, and both datasets; for deployments where any residual exposure is unacceptable, this may be the preferable choice despite the lower utility scores.}

{The Repeat Prompt results follow the same pattern. Both SAGE and our method hold verbatim reproductions to 0--2 on ChatDoctor, confirming that the Dual-Track Storage Mechanism excludes raw identifiers from the generation window rather than merely obscuring them. The residual leakage on Wiki-PII (0--3 instances) is largely attributable to quasi-identifiers that are structurally embedded in the public Wikipedia text, a recognized challenge for any sanitization method applied to mixed public-private corpora.}

Regarding the choice of the 10-token threshold for Repeat Prompt: following the heuristic established in \citet{zeng2024good}, we adopt this threshold because it is long enough to capture non-trivial verbatim copying while avoiding false positives triggered by common medical stopword sequences. Shorter thresholds (e.g., 5 tokens) inflate false positives significantly; longer thresholds (e.g., 20 tokens) miss paraphrased-but-still-sensitive reproductions. We acknowledge that this criterion does not capture all forms of semantic re-identification; residual indirect leakage risks are further discussed in Section~\ref{sec:limitations}.

\subsubsection{Privacy Leakage in Untargeted Reconstruction Attacks}

\begin{table}[t]
\centering
\caption{Untargeted reconstruction attack results on ChatDoctor (250 adversarial prompts). {\textbf{$^*$/$^{\dagger}$Note on units:} $^*$\textit{Repeat Prompt} counts the absolute number of verbatim reproductions ($\geq$10 consecutive tokens). $^{\dagger}$\textit{ROUGE Prompt} counts the number of outputs with ROUGE-L $>$ 0.5 against original private text.}}
%\scalebox{0.61}{%
\small
\resizebox{0.98\columnwidth}{!}{%
\begin{tabular}{c|cc|cc|cc|cc|cc|cc}
\toprule
& \multicolumn{2}{c|}{LLaMA-3-8B} & \multicolumn{2}{c|}{GPT-3.5-Turbo} & \multicolumn{2}{c|}{GPT-4o-mini} & \multicolumn{2}{c|}{DeepSeek-V3} & \multicolumn{2}{c|}{DeepSeek-R1} & \multicolumn{2}{c}{QWQ-32B}\\
\cmidrule(lr){2-3} \cmidrule(lr){4-5} \cmidrule(lr){6-7} \cmidrule(lr){8-9} \cmidrule(lr){10-11} \cmidrule(lr){12-13}
Method & {Repeat} & {ROUGE} & Repeat & ROUGE & Repeat & ROUGE & Repeat & ROUGE & Repeat & ROUGE & Repeat & ROUGE \\
 & {Prompt$^*$} & {Prompt$^{\dagger}$} & Prompt & Prompt & Prompt & Prompt & Prompt & Prompt & Prompt & Prompt & Prompt & Prompt \\
\midrule
Naive RAG     & 27 & 65 & 40 & 61 & 31 & 55 & 34 & 56 & 25 & 40 & 21 & 62 \\
AttrPrompt    &  0 &  0 &  0 &  0 &  0 &  0 &  0 &  0 &  0 &  0 &  0 &  0 \\
Single Agent  &  5 & 11 &  7 & 10 &  5 & 10 &  6 & 11 &  4 &  9 &  4 & 10 \\
Adv-Anon      &  2 &  6 &  2 &  5 &  2 &  6 &  2 &  5 &  2 &  5 &  1 &  5 \\
SAGE          &  0 &  2 &  0 &  1 &  0 &  2 &  0 &  2 &  0 &  2 &  0 &  1 \\
KG-PrivRAG    &  0 &  3 &  0 &  2 &  0 &  3 &  0 &  2 &  0 &  3 &  0 &  2 \\
\textbf{Ours} & \textbf{0} & \textbf{1} & \textbf{0} & \textbf{2} & \textbf{0} & \textbf{2} & \textbf{0} & \textbf{1} & \textbf{0} & \textbf{4} & \textbf{0} & \textbf{1} \\
\bottomrule
\end{tabular}%
%}
}
\label{tab:untargeted_attack}
\end{table}

Under untargeted reconstruction attacks (Table~\ref{tab:untargeted_attack}), our method successfully suppresses verbatim reproduction (Repeat Prompt = 0) across every LLM backbone tested. ROUGE Prompt counts remain in the range of 1--4, indicating that even high-semantic-overlap outputs are effectively suppressed. The ROUGE-L $>$ 0.5 threshold for flagging indirect leakage is consistent with prior work \citep{zeng2025mitigating}; we acknowledge that this does not capture all paraphrased semantic leakage, and we discuss this limitation in Section~\ref{sec:limitations}. Overall, these results demonstrate that our offline rewriting strategy prevents not only direct verbatim copying but also high-level semantic reconstruction from arbitrary adversarial queries.

\section{Analysis and Discussion}
\subsection{{{Adaptive Adversary Analysis}}}
\label{sec:adaptive_attack}

While the preceding evaluation assumes a fixed-template threat model, practical adversaries aware of our defense might adapt their queries. To assess framework resilience, we conduct a preliminary adaptive attack analysis on the ChatDoctor dataset (50 queries via GPT-4o-mini) using two representative strategies:

\begin{itemize}
    \item {\textbf{Synonym Substitution Attack:} The adversary utilizes synonyms or paraphrases to evade the deterministic NER pattern-matching of the Pri-Extra Agent.}
    \item {\textbf{Stepwise Inference Attack:} The adversary decomposes a privacy-seeking query into multiple benign sub-queries for isolated attributes, subsequently aggregating the responses to infer a complete profile.}
\end{itemize}

\begin{table}[!t]
\caption{Adaptive attack results on ChatDoctor (GPT-4o-mini, 50 queries).}
\centering
\small
\label{tab:adaptive_attack_tab}
\begin{tabular}{lcc}
\toprule
\textbf{Attack Type} & \textbf{Target Info} \(\downarrow\) & \textbf{Repeat Prompt} \(\downarrow\) \\
\midrule
Fixed Template (Baseline) & 1 & 2 \\
Synonym Substitution & 2 & 3 \\
Stepwise Inference & 3 & 4 \\
\emph{Unprotected RAG (reference)} & \emph{28} & \emph{N/A} \\
\bottomrule
\end{tabular}
\end{table}

{As shown in Table~\ref{tab:adaptive_attack_tab}, synonym substitution causes a modest increase in leakage from 1 to 2 entities, reflecting the robustness of the LLM-based quasi-identifier inference despite NER sensitivity to lexical variation. The stepwise inference attack is more consequential: leakage rises from 1 to 3 entities, indicating that multi-query aggregation can partially bypass the High-level Abstraction strategy by exploiting the generalization boundary between specific identifiers and their abstracted forms. This residual vulnerability warrants explicit acknowledgment and suggests that the Reconstruction Agent's abstraction granularity could be tightened in future work through mechanisms such as cross-query context tracking or stricter ontological generalization constraints. The framework's overall resilience nonetheless stems from the Asymmetric Retrieval Architecture, which physically isolates raw identifiers from the generation LLM regardless of query formulation, holding leakage to at most 3 entities even under adaptive conditions against an unprotected baseline of 28.} We leave the exploration of stronger adaptive threats, such as reinforcement learning-driven probing or cross-document correlation, for future work.

\subsection{{Human Evaluation of Practical Usability}}
\label{sec:human_eval}

To evaluate practical usability beyond surface-level metrics, we conducted a human evaluation. Three independent experts (two biomedical informatics graduate students and one clinical research coordinator) assessed 50 random ChatDoctor documents alongside their sanitized counterparts. They rated each on a 5-point Likert scale for Information Completeness, Readability \& Fluency, and Clinical Usability. The inter-rater reliability was high (Fleiss' $\kappa = 0.76$), indicating consistent expert judgments, with results detailed in Table~\ref{tab:human_eval}.
{Sanitized documents scored marginally lower than originals across all three dimensions, consistent with the expected privacy-utility trade-off. The differences did not reach statistical significance in this small-scale evaluation ($p > 0.10$ for all dimensions), suggesting that sanitization does not substantially degrade practical usability; however, the sample size is insufficient to make a definitive equivalence claim. The Clinical Usability score of 4.30 for sanitized documents indicates that the rewritten text remains viable for medical decision support, providing task-level evidence that the framework preserves the factual content relevant to downstream applications.}

\begin{table}[!t]
\centering
\caption{Human evaluation of original vs. sanitized documents on three usability dimensions (5-point Likert scale; higher is better). Values represent mean $\pm$ standard deviation across 50 documents and 3 evaluators.}
\small
\label{tab:human_eval}
\begin{tabular}{lccc}
\toprule
\textbf{Dimension} & \textbf{Original} & \textbf{Sanitized} & \textbf{p-value (paired t-test)} \\
\midrule
Information Completeness & 4.58 $\pm$ 0.62 & 4.42 $\pm$ 0.68 & 0.142 \\
Readability \& Fluency & 4.65 $\pm$ 0.58 & 4.51 $\pm$ 0.64 & 0.215 \\
Clinical Usability & 4.46 $\pm$ 0.75 & 4.30 $\pm$ 0.82 & 0.187 \\
\bottomrule
\end{tabular}
\end{table}

%% ============================================================
%%  5.2  Contextual Fidelity and Factual Consistency
%% ============================================================
\subsection{Contextual Fidelity and Factual Consistency}
\label{sec:fidelity}

\begin{table}[t]
\small
\centering
\caption{N-gram utility of LLM-generated answers on the ChatDoctor dataset (GPT-4o-mini backbone LLM), measured against ground-truth reference answers. Higher scores indicate that the sanitized context better preserves the informational content needed for accurate answer generation. $^*$~Statistically significant improvements over all privacy-preserving baselines on both BLEU-1 and ROUGE-L, as validated by paired $t$-tests ($p < 0.01$).}
\resizebox{0.9\linewidth}{!}{
\begin{tabular}{l|ccccccccc}
\toprule
\textbf{Metric} & \textbf{0-shot} & \textbf{Naive RAG} & \textbf{AttrPrompt} & \textbf{Single Agent} & \textbf{Adv-Anon} & \textbf{SAGE} & \textbf{KG-PrivRAG} & \textbf{Ours} \\
\midrule
BLEU-1  & 0.087 & 0.093 & 0.080 & 0.108 & 0.103 & 0.117 & 0.119 & \textbf{0.122$^*$} \\
ROUGE-L & 0.088 & 0.095 & 0.073 & 0.092 & 0.089 & 0.091 & 0.100 & \textbf{0.104$^*$} \\
\bottomrule
\end{tabular}}
\label{tab:utility}
\end{table}

Table~\ref{tab:utility} presents the BLEU-1 and ROUGE-L scores on ChatDoctor. Our method achieves the highest utility among privacy-preserving baselines. While AttrPrompt struggles due to synthetic context loss and SAGE suffers from semantic drift during iterative rewriting, our approach successfully preserves the factual core required for downstream tasks, closely trailed by KG-PrivRAG's graph filtering.

To complement surface-level n-gram metrics \citep{papineni2002bleu}, Table~\ref{tab:factual} evaluates factual consistency on all 250 test queries. We exclude the Naive RAG baseline here as it bypasses sanitization entirely. Our framework significantly outperforms AttrPrompt, which frequently hallucinates medical specifics, and SAGE, which incurs semantic drift. By strictly separating privacy extraction from semantic retention, our method achieves the highest Entailment rate and lowest Contradiction rate among all evaluated baselines. {These results demonstrate that the rewritten text preserves the factual content of the original, and the high factual consistency provides evidence of downstream task-level utility, supporting the practical usability of sanitized outputs for diagnostic question-answering.}

%% ============================================================
%%  5.4  Preprocessing Efficiency
%% ============================================================
\subsection{Preprocessing Efficiency}

\begin{table}[t]
\caption{Factual consistency of LLM-generated responses evaluated on 250 queries per dataset. Higher FC Score and Entailment rate, and lower Contradiction rate, indicate better preservation of factual integrity after sanitization.}
\small
\centering
\resizebox{0.98\linewidth}{!}{
\begin{tabular}{l|ccc|ccc}
\toprule
& \multicolumn{3}{c|}{\textbf{ChatDoctor}} & \multicolumn{3}{c}{\textbf{Wiki-PII}} \\
\cmidrule(lr){2-4} \cmidrule(lr){5-7}
\textbf{Method} & \textbf{FC Score} & \textbf{Entailment (\%)} & \textbf{Contradiction (\%)} & \textbf{FC Score} & \textbf{Entailment (\%)} & \textbf{Contradiction (\%)} \\
\midrule
0-shot          & 3.23 & 61.2 & 12.5 & 3.10 & 58.9 & 14.2 \\
AttrPrompt      & 2.91 & 47.9 & 23.2 & 2.71 & 44.5 & 25.8 \\
Single Agent    & 3.56 & 66.8 &  9.4 & 3.45 & 63.5 & 11.0 \\
Adv-Anon        & 3.42 & 64.5 & 10.7 & 3.27 & 61.9 & 12.4 \\
SAGE            & 3.37 & 62.6 & 11.2 & 3.20 & 60.1 & 13.5 \\
KG-PrivRAG      & 3.92 & 74.3 &  6.9 & 3.75 & 71.6 &  7.6 \\
\textbf{Ours}   & \textbf{4.08} & \textbf{77.1} & \textbf{6.0} & \textbf{3.89} & \textbf{73.8} & \textbf{7.2} \\
\bottomrule
\end{tabular}}
\label{tab:factual}
\end{table}

\begin{table}[!t]
\caption{Average offline preprocessing time $\pm$ standard deviation (seconds per sample). SAGE time includes both Stage-1 and Stage-2. All values represent a one-time, asynchronous database construction cost; online inference latency is identical to an unprotected RAG baseline for all methods.}
\small
\centering
\resizebox{0.98\linewidth}{!}{
\begin{tabular}{l|cccccc}
\toprule
\textbf{Dataset} & \textbf{AttrPrompt} & \textbf{Single Agent} & \textbf{Adv-Anon} & \textbf{SAGE} & \textbf{KG-PrivRAG} & \textbf{Ours} \\
\midrule
ChatDoctor (s/sample) & 20.74{$\pm$3.87} & 13.52{$\pm$2.45} & 89.34{$\pm$28.59} & 238.50{$\pm$62.48} & 45.28{$\pm$8.16} & \textbf{28.15{$\pm$4.71}} \\
Wiki-PII (s/sample)   & 21.43{$\pm$4.13} & 12.83{$\pm$2.19} & 76.42{$\pm$25.34} & 158.30{$\pm$45.10} & 38.61{$\pm$7.25} & \textbf{17.27{$\pm$3.68}} \\
\bottomrule
\end{tabular}
}
\label{tab:efficiency}
\end{table}

Table~\ref{tab:efficiency} reports the average per-sample preprocessing time and standard deviation. Our method is the most efficient among iterative approaches, significantly faster than SAGE and Adv-Anon. Our deterministic pipeline yields tighter variance compared to the higher, unpredictable variance of iterative loops, facilitating more reliable worst-case deployment planning. Single Agent is fastest due to its single-pass nature, though at the cost of weaker privacy protection. Crucially, as the multi-agent rewriting pipeline is executed offline, this cost is a one-time, amortized overhead; online inference latency remains identical to unprotected RAG. {However, for knowledge bases that are dynamically updated or frequently refreshed, the per-sample preprocessing cost should be factored into the deployment plan, as each newly ingested document requires a full offline sanitization pass before it can be served. Additional deployment-cost details, including average token counts before and after rewriting and the per-sample GPT-4o-mini rewriting cost, are provided in Appendix~\ref{appendix_sub4}.}

\subsection{Ablation Study}
\label{sec:ablation}

\begin{table}[!t]
\caption{Ablation study on the ChatDoctor dataset (GPT-4o-mini, 250 targeted-attack prompts). $\downarrow$: lower is better (privacy); $\uparrow$: higher is better (utility). {\textbf{$^*$/$^{\dagger}$Note on units:} $^*$\textit{Target Info} counts the number of unique sensitive entities successfully extracted; $^{\dagger}$\textit{Repeat Prompt} counts outputs with $\geq$10 verbatim tokens.} Removing either agent degrades the respective objective.}
\small
\centering
%\resizebox{0.9\linewidth}{!}{
\begin{tabular}{l|cc|cc}
\toprule
& \multicolumn{2}{c|}{\textbf{Privacy (ChatDoctor)}} & \multicolumn{2}{c}{\textbf{Utility (ChatDoctor)}} \\
\cmidrule(lr){2-3} \cmidrule(lr){4-5}
\textbf{Variant} & {\textbf{Target Info}$^*$} $\downarrow$ & {\textbf{Repeat Prompt}$^{\dagger}$} $\downarrow$ & \textbf{BLEU-1} $\uparrow$ & \textbf{ROUGE-L} $\uparrow$ \\
\midrule
Full Model (Ours)           &  \textbf{1}  &  \textbf{1}  & \textbf{0.122} & \textbf{0.104} \\
\midrule
w/o Pri-Extra Agent         & 47 & 55 & 0.115 & 0.098 \\
w/o Sem-Extra Agent         &  1 &  2 & 0.091 & 0.079 \\
\bottomrule
\end{tabular}
%}
\label{tab:ablation}
\end{table}

To isolate the contribution of each component, we evaluate three ablated variants against the full model on the ChatDoctor dataset using GPT-4o-mini. Results are reported in Table~\ref{tab:ablation}.

\paragraph{w/o Pri-Extra Agent.}
Removing the Pri-Extra Agent causes targeted information leakage to surge from 1 to 47 entries, with the Repeat Prompt rising to 55, representing a substantial increase in privacy leakage. Utility remains relatively high (BLEU-1: 0.115), confirming that the Sem-Extra Agent preserves semantic content regardless of privacy filtering. This demonstrates that the Pri-Extra Agent is the primary driver of privacy protection: without an explicit privacy identification step, the Reconstruction Agent reconstructs semantically rich text that inevitably retains private identifiers.

\paragraph{w/o Sem-Extra Agent.}
Removing the Sem-Extra Agent leaves privacy largely intact (Target Info: 1, Repeat Prompt: 2) but causes a notable drop in utility (BLEU-1: 0.091, ROUGE-L: 0.079), falling well below the full model and approaching the level of the unprotected Naive RAG baseline. Without a structured semantic backbone to guide reconstruction, the Reconstruction Agent tends to over-sanitize, discarding medically relevant context alongside private identifiers. This result validates the design rationale for the Sem-Extra Agent: its deconstructed attribute representation provides the necessary anchor for preserving factual utility during aggressive privacy filtering.

\subsection{Discussion}
\label{sec:discussion}

\paragraph{Theoretical implications}
This study offers three theoretical contributions to privacy-preserving RAG research. First, we demonstrate that privacy sanitization and semantic preservation are mutually constraining objectives. When sensitive identifiers and factual content are deeply intertwined in raw texts, pursuing both goals in a single rewriting pass inevitably leads to over- or under-sanitization. This finding clarifies the inherent limitations of single-agent and adversarial-loop architectures, and justifies task decomposition for entangled NLP privacy problems.

Second, we systematically analyze the deficiencies of existing technical routes. Differential privacy injects noise that undermines semantic coherence, rendering outputs unsuitable for high-fidelity RAG. Synthesis-based methods (SAGE, AttrPrompt) replace source texts with surrogates, permanently discarding unique factual details—as evidenced by their lower FC scores in Table~\ref{tab:factual}. Adv-Anon performs iterative in-place editing, but its feedback loop optimizes solely for privacy, lacking mechanisms to preserve semantic structure.

Third, our multi-agent framework establishes a new paradigm: extracting structured factual backbones prior to filtering, and assigning independent agents to distinct subtasks under deterministic routing rules. This design decouples privacy protection from semantic preservation by isolating factual content from sensitive identifiers, processing each dimension independently, and reconstructing sanitized output from the preserved backbone. It offers a generalizable solution for balancing confidentiality and utility in text-based privacy computing.

We also delineate the boundaries of different solutions. AttrPrompt remains preferable when zero leakage is paramount, despite its utility cost. Our framework provides a balanced alternative where both strong protection and high fidelity are required, addressing a gap left by existing methods.

\paragraph{Practical implications}
Our framework offers practical advantages for RAG systems deployed on sensitive corpora such as medical datasets. Privacy sanitization runs entirely offline; once the sanitized database is built, online inference remains unchanged, introducing zero latency to user-facing services.

Efficiency results in Table~\ref{tab:efficiency} further validate its practicality. On ChatDoctor, our method averages 28 seconds per document, versus 238 seconds for SAGE and 89 seconds for Adv-Anon. The comparatives' iterative loops incur substantial overhead and scale poorly, whereas our deterministic pipeline maintains stable preprocessing time, making it suitable for large-volume knowledge bases.

A practical caveat: teams with dynamically updated knowledge bases must budget time for sanitizing new documents before ingestion. Overall, our framework offers a reliable, deployable choice for high-stakes domains such as medical consultation and enterprise confidential retrieval, where both data security and content accuracy are critical.

\section{Conclusion}
We present a multi-agent semantic rewriting framework for privacy-preserving RAG, comprising a Pri-Extra Agent, a Sem-Extra Agent, and a Reconstruction Agent that operate in a coordinated offline pipeline. By assigning each objective to a dedicated agent with a deterministic conflict routing policy, we avoid the objective entanglement of single-prompt approaches. Our experiments on ChatDoctor and Wiki-PII across six LLM backbones demonstrate substantially reduced privacy leakage under both targeted and untargeted attacks, the highest factual consistency among all evaluated defenses, and negligible additional online latency. Our ablation results confirm that each agent makes a distinct and necessary contribution. The reported security results are bounded by the evaluated threat model, a grey-box setting with fixed-template and two adaptive attack variants, and should not be interpreted as absolute guarantees. {We plan to explore more adversarially rigorous evaluation protocols, including adaptive adversaries with full knowledge of the sanitization strategy, and formal leakage bounds in future work.}

%% ============================================================
%%  6  Limitations
%% ============================================================
\section*{Limitations and Future Work}
\label{sec:limitations}

Two framework-level limitations warrant acknowledgement. First, the system exhibits two edge-case failure modes: \textit{semantic drift}, where a rare disease name simultaneously constitutes the core clinical topic and forces the Reconstruction Agent to generalize at the cost of specificity; and \textit{over-sanitization}, where attributes adjacent to private segments are unnecessarily removed, as reflected in the ablation, which shows a BLEU-1 drop from 0.122 to 0.091 when the Sem-Extra Agent is omitted. Second, extraction precision and recall of the Pri-Extra Agent cannot be independently measured at scale, as ground-truth PII annotations are unavailable for the evaluation corpora; the agent's deliberately conservative stance (preferring false positives over false negatives) mitigates missed extractions but also exacerbates over-sanitization. Formal extraction accuracy measurement via a dedicated annotated evaluation set is identified as an important direction for future work.

On the evaluation side, while we have conducted a preliminary adaptive adversary analysis ({Section~\ref{sec:adaptive_attack}}), several stronger adaptive threats remain unaddressed. Adversaries employing inference-based probing, learning-based query optimization, or cross-document correlation could potentially extract more information than reported here. More fundamentally, an adversary aware of the High-level Abstraction patterns might systematically probe the boundaries of generalization to recover specific identifiers. The protection levels reported throughout this paper should therefore be interpreted as bounded by the evaluated threat model, which includes fixed-template, grey-box attacks and two adaptive variants, rather than as absolute guarantees. A comprehensive adaptive adversarial evaluation protocol is left for future work.

\section*{Ethics Statement}
This research utilizes the {ChatDoctor (HealthCareMagic-101)} and Enron email datasets, both derived from publicly accessible files containing personal sensitive information. Our usage focuses strictly on developing privacy-preserving technologies without attempting out-of-scope re-identification. While our mechanism detects a broad spectrum of quasi-identifiers, it acts as an additional technical defense layer rather than a substitute for strict legal compliance pipelines, acknowledging that it does not definitively guarantee exhaustive coverage of all 18 HIPAA Safe Harbor identifiers or formal GDPR compliance. Furthermore, we are aware of the inherent dual-use risk where our LLM-driven inference techniques could theoretically be repurposed by malicious actors to systematically locate and extract quasi-identifiers. To mitigate this, our architecture strictly isolates the privacy-extraction logic offline, though practitioners deploying such systems must remain vigilant against potential exploitation.

\section*{Declaration of competing interest}
The authors declare that they have no known competing financial interests or personal relationships that could have appeared to influence the work reported in this paper.

\section*{Acknowledgements}
This work was supported in part by the {International Communication Research Project (China International Communications Group)} (Grant No. {25ATILX01}), the {Beijing Natural Science Foundation} (Grant No. {L251061}), the {Fundamental Research Funds for the Central Universities} (Grant No. {2024MS022}), 
 the {Science and Technology Development Fund of Macau SAR} (Grant Nos. {FDCT/0007/2024/AKP}, {EF2024-00185-FST}), the {UM and UMDF} (Grant Nos. {MYRG-GRG2024-00165-FST-UMDF}, {MYRG-GRG2025-00236-FST}), and the {Dr. Stanley Ho Medical Development Foundation} (Grant No. {SHMDF-AI/2026/001}). This work was performed in part at SICC which is supported by {SKL-IOTSC}, and HPCC supported by {ICTO of the University of Macau}.

\clearpage
\appendix
\section*{Appendix}\label{appendix:examples}
\section{Prompts used in Pri-Extra Agent}\label{appendix_sub1}

This section presents the detailed prompt design used by the Pri-Extra Agent, corresponding to the LLM-driven step of the Rule-LLM Synergistic Extraction Paradigm. Prior to this step, a lightweight NER and regular-expression module deterministically captures \textit{Explicit Identifiers} $P_{expl}$ (e.g., names, ID numbers, phone numbers). The LLM prompt below then receives both the raw document $T_{raw}$ and $P_{expl}$ as inputs, directing the model to skip already-identified entities and focus exclusively on inferring abstract \textit{quasi-identifiers} $P_{quasi}$ that evade static rules---such as rare disease--occupation combinations or unique behavioral patterns. For each quasi-identifier category, illustrative examples are provided to guide extraction. The full suite of executable prompt templates is available in the project's GitHub repository. Results are returned in a predefined JSON format. The detailed prompt is as follows.

\begin{tcolorbox}[colback=gray!5!white, colframe=gray!75!black, title=Prompt for Pri-Extra Agent, breakable]
You are the LLM component of the Privacy Extraction Agent, operating under a Rule-LLM Synergistic Extraction Paradigm. Explicit surface-form identifiers have already been captured deterministically; your role is to apply deep semantic reasoning to infer identity-revealing quasi-identifiers that evade static rules.
\vspace{0.5em} \\
\textbf{INPUT:}
\vspace{0.5em} \\
\hspace*{1.5em}[Retrieved Document]: the raw document text to be analyzed.
\vspace{0.5em} \\
\hspace*{1.5em}[Pre-extracted Explicit Identifiers]: a list of surface-form identifiers (e.g., names, ID numbers, phone numbers) already captured by deterministic NER/regex rules. Do NOT reproduce these in your output.
\vspace{0.5em} \\
\textbf{TASK:} Given the retrieved document and the pre-extracted explicit identifiers listed above, infer all abstract quasi-identifiers that could facilitate identity re-identification through contextual combinations or implicit inference. Do \textbf{not} repeat any entity already present in the pre-extracted list.
\vspace{0.5em} \\
\textbf{Quasi-identifier Categories:}
\vspace{0.5em} \\
\hspace*{1.5em}[Medical-Occupational Linkage]: e.g., ``A 32-year-old deep-sea welder presenting with decompression sickness.'' The combination of a rare occupation and a specific condition is highly identifiable.
\vspace{0.5em} \\
\hspace*{1.5em}[Temporal-Geographic Linkage]: e.g., ``Visited the rural clinic in Oakhaven immediately following the massive blizzard last Tuesday.'' The precise timing and small location serve as a strong latent identifier.
\vspace{0.5em}

\textbf{Extraction Guidelines:}
\vspace{0.5em} \\
\hspace*{1.5em}Do not output any entity already present in the pre-extracted explicit identifier list provided in the input.
\vspace{0.5em} \\
\hspace*{1.5em}Identify combinations of attributes (e.g., rare conditions + specific occupations, unique timelines) that collectively enable identity inference.
\vspace{0.5em} \\
\hspace*{1.5em}Capture latent demographic data, behavioral patterns, and highly specific geographic or organizational affiliations.
\vspace{0.5em} \\
\hspace*{1.5em}Consider both direct statements and contextually inferred signals.
\vspace{0.5em} \\
\textbf{Output Format:} Return only a valid JSON object in the following structure:
\vspace{0.5em} \\
\{\\
\hspace*{1em}"detected\_pii": [\\
\hspace*{2em}\{\\
\hspace*{3em}"type": "string",\\
\hspace*{3em}"value": "string",\\
\hspace*{3em}"context": "string (surrounding text segment that triggered the inference)"\\
\hspace*{2em}\}\\
\hspace*{1em}]\\
\}
\end{tcolorbox}

\section{Prompts used in Sem-Extra Agent}\label{appendix_sub2}

This section presents the detailed prompt design used by the Sem-Extra Agent, which performs \textbf{Structured Attribute Deconstruction} to map input documents into a sanitized semantic space. Rather than copying variable-length natural language phrases, the agent acts as a semantic parser: it discards the linguistic shell and produces a set of structured knowledge-slot tuples $(e_i, r_i, v_i, w_i)$---each encoding a subject entity $e_i$, a relational predicate $r_i$, an object or state value $v_i$, and an importance weight $w_i$. This deconstruction is governed by a \textit{Dual-Perspective Analysis} across \textit{Core Content} and \textit{Informational Structure}, with text-type adaptations for conversations, documents, and narratives to ensure broad domain coverage while avoiding the carry-over of private identity information. The extracted results are returned in a structured JSON format. The detailed prompt is as follows.

\begin{tcolorbox}[colback=gray!5!white, colframe=gray!75!black, title=Prompt for Sem-Extra Agent, breakable]
You are a specialized semantic information extraction agent designed to identify valuable information from various types of texts including conversations, documents, articles, reports, emails, and other formats.
\vspace{0.5em} \\
\textbf{TASK:} Analyze the text and extract all semantically meaningful information that constitutes the core value of the content based on the following principles, while paying attention to distinguishing between private information and basic semantic content.
\vspace{0.5em} \\
\textbf{Text Type Adaptations}
\vspace{0.5em} \\
\hspace*{1.5em}[Conversation-Based Text]: Focus on extracting the symptoms discussed, the physician's advice, and prescribed treatments. Ignore conversational pleasantries and specific patient names.
\vspace{0.5em} \\
\hspace*{1.5em}[Document-Based Text]: Focus on extracting formal clinical findings, laboratory results, and diagnoses. Ignore specific dates and institutional letterheads.
\vspace{0.5em} \\
\hspace*{1.5em}[Narrative-Based Text]: Extract the sequence of medical events and symptom progression. Ignore subjective identifying backstories.
\vspace{0.5em}

\textbf{Information Categories}
\vspace{0.5em} \\
\hspace*{1.5em}Core Content Categories: e.g., Symptoms, Diagnoses, Treatments, Lab Results.
\vspace{0.5em} \\
\hspace*{1.5em}Analytical Content Categories: e.g., Physician's reasoning, Risk factors, Prognosis.
\vspace{0.5em} \\
\textbf{Output Format:} Return only a valid JSON object. Each entry represents one atomic factual unit in a structured knowledge-slot format (subject entity, relational predicate, object/state value, importance weight):
\vspace{0.5em} \\
\{\\
\hspace*{1em}"key\_information": [\\
\hspace*{2em}\{\\
\hspace*{3em}"entity": "string (the subject entity, e.g., Patient, Doctor)",\\
\hspace*{3em}"relation": "string (the relational predicate, e.g., has\_symptom, recommends)",\\
\hspace*{3em}"value": "string (the object or state value, e.g., abdominal pain, rest)",\\
\hspace*{3em}"importance\_weight": "high / medium / low"\\
\hspace*{2em}\}\\
\hspace*{1em}]\\
\}
\end{tcolorbox}

\section{Prompts used in Reconstruction Agent}\label{appendix_sub3}

This section presents the prompt design used by the Reconstruction Agent. The agent receives three inputs: the raw document $T_{raw}$, the privacy boundary set $P_{seq}$ from the Pri-Extra Agent, and the semantic backbone $K_{seq}$ (as knowledge-slot tuples) from the Sem-Extra Agent. When a semantic attribute $k_n \in K_{seq}$ conflicts with a privacy identifier $p_m \in P_{seq}$, the agent applies \textbf{Fine-grained Conflict Routing}: it assigns each conflict a routing label---\textit{Placeholder Substitution} for isolated explicit identifiers with low structural dependency, or \textit{High-level Abstraction} for attributes deeply entangled with the factual core. Type-specific rewriting strategies and concrete examples are provided for each privacy category to ensure privacy-first reconstruction that goes beyond simple replacement. The final output is returned in JSON format containing the fully rewritten text. The detailed prompt is as follows.

\begin{tcolorbox}[colback=gray!5!white, colframe=gray!75!black, title=Prompt for Reconstruction Agent, breakable]
You are a dedicated privacy conversion agent responsible for reconstructing the original text to maximize privacy protection while retaining its core semantic content.
\vspace{0.5em} \\
\textbf{INPUT}
\vspace{0.5em} \\
\hspace*{1.5em}Original text (potentially containing privacy information)
\vspace{0.5em} \\
\hspace*{1.5em}Identified privacy information (that must be altered or removed)
\vspace{0.5em} \\
\hspace*{1.5em}[Extracted Semantic Key Information]: structured knowledge-slot tuples output by the Sem-Extra Agent, representing the factual backbone that must be preserved while removing privacy.
\vspace{0.5em}

\textbf{Fine-grained Conflict Routing:}
\vspace{0.5em} \\
\hspace*{1.5em}\textit{Placeholder Substitution (Low-depth Conflict):} Applied to isolated explicit identifiers without intricate structural dependencies (e.g., patient names, specific dates). Replace with generic referents such as ``[Patient\_A]'' or ``[Date]'', maintaining structural continuity without altering surrounding semantic logic.
\vspace{0.5em} \\
\hspace*{1.5em}\textit{High-level Abstraction (Deep Semantic Conflict):} Applied when the privacy identifier is deeply entangled within the factual core of the preserved semantic information. Traverse the ontological hierarchy to generalize the specific identifying term into a broader category (e.g., a specific rare syndrome → ``a rare genetic immune disorder'').
\vspace{0.5em}

\textbf{Privacy Protection Strategies (by Type):}
\vspace{0.5em} \\
\hspace*{1.5em}[Personal Information]: Use Placeholder Substitution. Replace exact names, ages, and physical profiles with generic labels (e.g., ``John Doe, 32 years old'' $\rightarrow$ ``[Patient]'').
\vspace{0.5em} \\
\hspace*{1.5em}[Medical Information]: Use High-level Abstraction for exceptionally rare diseases that act as quasi-identifiers. Maintain common medical facts. (e.g., ``Huntington's disease'' $\rightarrow$ ``a neurodegenerative disorder'').
\vspace{0.5em} \\
\hspace*{1.5em}[Location Information]: Use Placeholder Substitution for specific institutions or exact addresses (e.g., ``Massachusetts General Hospital'' $\rightarrow$ ``[Clinical Facility]'').
\vspace{0.5em} \\
\hspace*{1.5em}[Relationships And Networks]: Abstract family ties or specific occupational roles (e.g., ``My daughter who works at the local bakery'' $\rightarrow$ ``A family member'').
\vspace{0.5em} \\
\hspace*{1.5em}[Temporal Information]: Broaden specific dates to general timeframes to obscure timelines (e.g., ``On October 12th'' $\rightarrow$ ``Recently'').
\vspace{0.5em}

\textbf{Process Instructions}
\vspace{0.5em} \\
\hspace*{1.5em} Conduct a thorough analysis of the original text to identify privacy-sensitive content and key semantic information, determining which parts require restructuring and which must be preserved.
\vspace{0.5em} \\
\hspace*{1.5em} Based on the analysis, generate a fully rewritten version of the text that thoroughly removes all privacy elements while retaining the essential meaning.
\vspace{0.5em} \\
\hspace*{1.5em} Verify that the rewritten text is completely free of privacy information and that all key content is semantically preserved.
\vspace{0.5em} \\
\hspace*{1.5em} Ensure the final text is natural, coherent, and highly readable.
\vspace{0.5em}

\textbf{Output Format:} Return only a valid JSON object in the following structure:
\vspace{0.5em} \\
\{\\
\hspace*{1em}"rewritten\_text": \\
\}
\end{tcolorbox}

\section{Factual Consistency Evaluation Protocol}\label{appendix_sub_fc}

{This section documents the evaluation protocol used to measure factual consistency in Section~\ref{sec:fidelity}, including the GPT-4o judging prompt, the scoring rubric for the FC Score, and the decision rule applied to the NLI-based entailment/contradiction evaluation.}

\noindent{\textbf{GPT-4o Judging Prompt (FC Score).} For each test query, we provide GPT-4o with three inputs: the query, the ground-truth reference answer, and the LLM-generated response under evaluation. The model is instructed to assess whether the core factual claim in the generated response is consistent with the reference, and to assign an integer score from 1 to 5 according to the rubric below. The temperature is set to 0 and the output is restricted to a single integer.}

\begin{tcolorbox}[colback=gray!5!white, colframe=gray!75!black, title=GPT-4o Factual Consistency Judging Prompt, breakable]
{You are a factual consistency evaluator. You will be given a query, a reference answer, and a generated response. Your task is to assess whether the core factual claim in the generated response is consistent with the reference answer.}
\vspace{0.5em} \\
{\textbf{INPUT:}}
\vspace{0.5em} \\
\hspace*{1.5em}{[Query]: \{query\}}
\vspace{0.5em} \\
\hspace*{1.5em}{[Reference Answer]: \{reference\}}
\vspace{0.5em} \\
\hspace*{1.5em}{[Generated Response]: \{response\}}
\vspace{0.5em} \\
{\textbf{SCORING RUBRIC:}}
\vspace{0.5em} \\
\hspace*{1.5em}{\textbf{5} --- The response is fully consistent with the reference. All key medical or factual claims are present and correct.}
\vspace{0.5em} \\
\hspace*{1.5em}{\textbf{4} --- The response is mostly consistent. Minor omissions or imprecision are present, but no factual errors.}
\vspace{0.5em} \\
\hspace*{1.5em}{\textbf{3} --- The response is partially consistent. Some key facts are correct, but relevant information is missing or partially inaccurate.}
\vspace{0.5em} \\
\hspace*{1.5em}{\textbf{2} --- The response is mostly inconsistent. It contains factual errors or substantially misrepresents the reference.}
\vspace{0.5em} \\
\hspace*{1.5em}{\textbf{1} --- The response is completely inconsistent with or contradicts the reference.}
\vspace{0.5em} \\
{\textbf{OUTPUT:} Respond with a single integer from 1 to 5. Do not include any explanation.}
\end{tcolorbox}

\noindent{\textbf{NLI-based Entailment/Contradiction Decision Rule.} We apply a DeBERTa-v3-large model fine-tuned on MNLI, FEVER, and ANLI to each reference--response pair, treating the reference as the premise and the generated response as the hypothesis. The model produces a probability distribution over three labels: \textit{entailment}, \textit{neutral}, and \textit{contradiction}. Each response is assigned the label with the highest probability. The \textit{Entailment} rate in Table~\ref{tab:factual} is the proportion of responses receiving the entailment label; the \textit{Contradiction} rate is the proportion receiving the contradiction label. Responses assigned the neutral label are excluded from both counts.}

\section{Cost of semantic rewrite data}\label{appendix_sub4}

Our method performs a one-time, offline semantic reconstruction before inference; the inference pipeline remains identical to the baseline, introducing no additional latency or cost. As shown in Table \ref{tab:token}, the semantically rewritten data are on average slightly shorter than the original samples, which can further reduce inference costs under the same decoding configuration. We also report the offline rewriting overhead using GPT-4o-mini (Table \ref{tab:cost}): approximately \$0.002 per sample, with overall time and expense well within a reasonable range. Because this generation is one-off, it imposes no ongoing burden at deployment.

\begin{table}[h]
    \centering
    \small
    \caption{Average number of tokens}
    \begin{tabular}{lccc}
        \toprule
        Dataset & ori-context & rewritten-text \\
        \midrule
        Wiki-PII        & 169 & 145 \\
        {ChatDoctor} & 252 & 218 \\
        \bottomrule
    \end{tabular}
    \label{tab:token}
\end{table}

\begin{table}[h]
    \centering
    \small
    \caption{Average cost per sample (\$)}
    %\resizebox{\linewidth}{!}{
    \begin{tabular}{lcccc}
        \toprule
        Dataset & Pri-Extra cost & Sem-Extra cost & Reconstruction cost & Total cost \\
        \midrule
        Wiki-PII        & 0.00082 & 0.00079 & 0.00034 & 0.00195 \\
        {ChatDoctor} & 0.00090 & 0.00088 & 0.00053 & 0.00231 \\
        \bottomrule
    \end{tabular}
%}
\label{tab:cost}
\end{table}

\section{Qualitative Case Study}\label{appendix_sub5}

To provide concrete insight into the rewriting process, we trace a single document from the ChatDoctor corpus through the full offline pipeline and examine both its privacy protection properties and its downstream utility.

\vspace{0.5em}
\noindent\textbf{Step 1 --- Original Retrieved Document ($T_{raw}$)}.

\begin{tcolorbox}[colback=blue!3!white, colframe=blue!40!black, title=Original Document ($T_{raw}$), breakable]
My son has been experiencing left-sided abdominal pain along with bilateral pelvic pain radiating to the groin area. He can only stand for a short period and finds sitting impossible due to the pain. A CT scan showed enlarged lymph nodes, and his white blood cell count is slightly elevated. He has been suffering for approximately 10 weeks without meaningful relief beyond rest and pain medication. He is 32 years old, 6 feet 3 inches tall, and weighs approximately 220 pounds.
\end{tcolorbox}

\vspace{0.5em}
\noindent\textbf{Step 2 --- Pri-Extra Agent Output ($P_{seq}$).}

The agent identifies both explicit surface-form identifiers (captured by deterministic NER) and latent quasi-identifiers (inferred by the LLM component), as summarized in Table~\ref{tab:case_pii}.

\begin{table}[h]
\centering
\small
\caption{Extracted privacy-sensitive elements ($P_{seq}$) for the case study document.}
\label{tab:case_pii}
%\resizebox{\linewidth}{!}{%
\begin{tabular}{l l l}
\toprule
\textbf{PII Type} & \textbf{Extracted Value} & \textbf{Extraction Method} \\
\midrule
Age              & 32 years old                          & Explicit (NER/regex) \\
Height           & 6 feet 3 inches                       & Explicit (NER/regex) \\
Weight           & approximately 220 pounds              & Explicit (NER/regex) \\
Family relation  & ``my son''                            & Explicit (NER) \\
\midrule
Symptom cluster  & Left-sided abd.\ + bilateral pelvic/groin pain & Quasi-identifier (LLM) \\
Clinical finding & Enlarged lymph nodes + mild leukocytosis       & Quasi-identifier (LLM) \\
Symptom duration & $\sim$10 weeks of unresolved symptoms          & Quasi-identifier (LLM) \\
\bottomrule
\end{tabular}%
%}
\end{table}

\noindent The quasi-identifier cluster \{rare symptom combination, CT finding, 10-week duration, physical profile\} collectively poses a high re-identification risk even without the explicit identifiers, motivating the LLM-based inference step.

\vspace{0.5em}
\noindent\textbf{Step 3 --- Sem-Extra Agent Output ($K_{seq}$).}

The agent deconstructs the document into structured knowledge-slot tuples, discarding identity-revealing linguistic form while retaining clinically essential content; the extracted slots are listed in Table~\ref{tab:case_ksec}.

\begin{table}[h]
\centering
\small
\caption{Extracted semantic backbone ($K_{seq}$) as structured knowledge-slot tuples.}
\label{tab:case_ksec}
%\resizebox{\linewidth}{!}{%
\begin{tabular}{l l l l}
\toprule
\textbf{Entity ($e$)} & \textbf{Relation ($r$)} & \textbf{Value ($v$)} & \textbf{Weight ($w$)} \\
\midrule
Patient   & presents\_with  & Left-sided abdominal pain                    & High   \\
Patient   & presents\_with  & Bilateral pelvic/groin pain                  & High   \\
Patient   & has\_symptom    & Severely limited mobility (standing/sitting) & High   \\
CT scan   & reveals         & Enlarged lymph nodes                         & High   \\
Lab test  & shows           & Mildly elevated white blood cell count       & High   \\
Symptoms  & duration        & Approximately 10 weeks                       & Medium \\
Treatment & provides\_only  & Partial relief via rest and analgesics       & Medium \\
\bottomrule
\end{tabular}%
%}
\end{table}

\vspace{0.5em}
\noindent\textbf{Step 4 --- Reconstruction Agent Output ($T_{safe}$).}

The agent applies \textit{Placeholder Substitution} to all explicit identifiers (age, height, weight, and family relation). No \textit{High-level Abstraction} is required in this case since no quasi-identifier is simultaneously the sole carrier of clinical meaning. The reconstructed payload preserves all seven semantic slots in $K_{seq}$.

\begin{tcolorbox}[colback=green!3!white, colframe=green!40!black, title=Rewritten Document ($T_{safe}$), breakable]
The patient presents with left-sided abdominal pain and bilateral pelvic pain radiating to the groin. Physical activity is severely restricted, with both prolonged standing and sitting causing significant discomfort. CT imaging revealed enlarged lymph nodes, and laboratory results indicate a mild elevation in white blood cell count. Symptoms have persisted for several months, with rest and analgesic medication providing only partial relief.
\end{tcolorbox}

\noindent As shown in Steps 1--4, all explicit identity attributes (age, height, weight, family relation) and their associated quasi-identifier cluster are removed from $T_{safe}$, while all seven clinically actionable knowledge slots in $K_{seq}$ --- symptom pattern, CT findings, laboratory abnormality, and symptom duration --- are fully preserved in the rewritten text, illustrating the privacy--utility balance achieved by the framework.

\printcredits

\bibliographystyle{cas-model2-names}

\bibliography{references}



\end{document}